# Brain Tissue Segmentation Using NeuroNet With Different Pre-processing Techniques

Fakrul Islam Tushar, Basel Alyafi, Md. Kamrul Hasan, Lavsen Dahal
Erasmus+ Joint Master In Medical Imaging and Applications, University of Girona, Spain
Email: f.i.tushar.eee@gmail.com, basel931991@gmail.com, kamruleeekuet@gmail.com, er.lavsen@gmail.com

*Abstract*—Automatic segmentation of brain Magnetic Resonance Imaging (MRI) images is one of the vital steps for quantitative analysis of brain for further inspection. In this paper, NeuroNet has been adopted to segment the brain tissues (white matter (WM), grey matter (GM) and cerebrospinal fluid (CSF)) which uses Residual Network (ResNet) in encoder and Fully Convolution Network (FCN) in the decoder. To achieve the best performance, various hyper-parameters have been tuned, while, network parameters (kernel and bias) were initialized using the NeuroNet pre-trained model. Different pre-processing pipelines have also been introduced to get a robust trained model. The model has been trained and tested on IBSR18 data-set. To validate the research outcome, performance was measured quantitatively using Dice Similarity Coefficient (DSC) and is reported on average as 0.84 for CSF, 0.94 for GM, and 0.94 for WM. The outcome of the research indicates that for the IBSR18 data-set, pre-processing and proper tuning of hyper-parameters for NeuroNet model have improvement in DSC for the brain tissue segmentation.

## I. INTRODUCTION

Brain image segmentation plays a crucial role in brain image analysis which extracts brain tissues, WM, GM, and CSF from a brain image by partitioning it into a set of disjoint regions. Pixels inside each of those regions should be homogeneous in space and intensity [1]. Segmentation of brain tissue helps to detect diseases like brain tumor, Alzheimers (AD), Parkinsons, dementia, schizophrenia, and traumatic injury [2]. The performance of brain tissue segmentation methods depends on several factors such as regions location, size, shape, texture, and contrast, especially when using multiple acquisition modalities with different properties [3].

Traditional brain tissue segmentation approaches can be classified into region-based, clustering-based, statistical based methods [1], like gaussian mixture models and atlas, and classification methods where features are finely engineered to be classified eventually using Support Vector Machines, for instance. In the first three algorithms, pipelines are usually used where a failure in one step might dramatically affect the final result. On the other hand, the fourth class suffers from hard feature selection and integration. Recently, Convolutional Neural Networks (CNNs) gained recognition in brain tissue segmentation [4]. Here, end-to-end segmentation is done by extracting features from images in an objective-oriented manner.

A recent and effective work by [5] has shown that Neuronet is a powerful network that can be used to segment raw brain images and outperformed previous tools. The extra feature is that this model can reproduce the output of multiple state-of-the-art tools simultaneously.

Our contribution in this paper is the use of two different pre-processing strategies followed by training with NeuroNet model to segment brain tissues for IBSR18 dataset and compare the performance with Dice Coefficient Similarity as a metric. These two pre-processing strategies are put under the microscope with different complexity levels, a simple one with standardization only against a more complicated one with registration, histogram equalization, then histogram matching. Then, a quantitative evaluation for the outcomes shows which strategy has been the most effective.

The remaining parts of the paper is organized as follows: Section II is dedicated to describe the dataset used, while, section III is used to explain the pre-processing pipelines. We illustrate the proposed method in section IV. We describe the implementation in section V and, experiment and results in section VI. We finally conclude in section VII.

## II. DATASET

The dataset used for this work, IBSR18, is a publicly-available dataset by the Center for Morphometric Analysis at Massachusetts General Hospital [6]. The dataset is composed of 18 T1-W volumes with different slice thicknesses. The image volumes used are skull stripped and bias field corrected. For this work, the dataset was divided randomly into three subsets: ten for training, five for validation, and three for testing. A brief description of the datast is given in Table I. The training subset was used for training the model, while, the validation subset was used to tune the proposed model. While, the test subset was eventually to evaluate the model. Fig. 1 shows the volume IBSR_01 and corresponding labels.

## III. PRE-PROCESING

The task of segmenting brain tissues become more challenging as different scanners are used with different parameters during acquisition. That usually leads to intensity heterogeneity, contrast variations, and different types of noise [7]. Therefore, data homogenization is often necessary.

TABLE I
SUMMARY ON IBSR18 DATASET USED IN THIS PAPER.

| Training Subset | | |
|---|---|---|
| Volume Name | Volume | Spacing (mm) |
| IBSR 01, 03, 04, 06 | 256 × 128 × 256 | 0.94 ×1.5 × 0.94 |
| IBSR 07,08,09 | 256 × 128 × 256 | 1 ×1.5 × 1 |
| IBSR 16,18 | 256 × 128 × 256 | 0.84 ×1.5 × 0.84 |
| Validation Subset | | |
| Volume Name | Volume | Spacing (mm) |
| IBSR 11,12 | 256 × 128 × 256 | 1 ×1.5 × 1 |
| IBSR 13,14 | 256 × 128 × 256 | 0.94 ×1.5 × 0.94 |
| IBSR 17 | 256 × 128 × 256 | 0.84 ×1.5 × 0.84 |
| Test Subset | | |
| Volume Name | Volume | Spacing (mm) |
| IBSR 02 | 256 × 128 × 256 | 0.94 ×1.5 × 0.9375 |
| IBSR 10 | 256 × 128 × 256 | 1 ×1.5 × 1 |
| IBSR 15 | 256 × 128 × 256 | 0.84 ×1.5 × 0.84 |

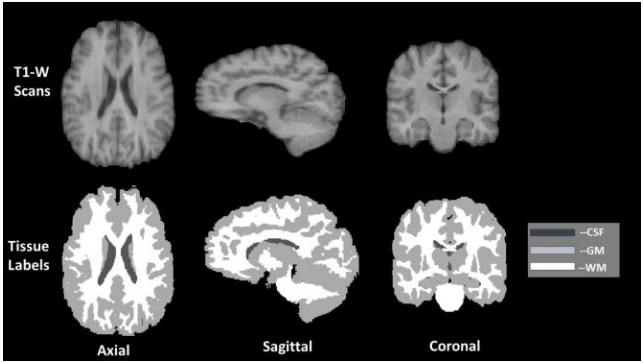

Fig. 1. Graphical description of IBSR 01 volume and corresponding ground truth labels in axial, sagittal and coronal view.

Authors in [8] used intra-subject registration as the first step of pre-processing for Lupus segmentation. Dolz et al. in [9] used volume-wise intensity normalization, bias field correction and skull-stripping. Shakeri et al. [10] used normalization for subcortical brain structure segmentation. Nyul et al. proposed a method consist of a training stage to find standard parameters then matching the histograms to a standard histogram through a transformation stage [11].

In this work, two different pre-processing pipelines were implemented to see the effect on the performance of the deep CNN. The idea is to apply a basic standardization pipeline first and compare the preformance with a second registration-based pipeline. Fig. 2 shows an overview of both pipelines.

### A. Pre-processing Pipeline-1

The input volumes were standardized to zero mean and unit standard deviation using the volume statistics as mentioned in the reference paper of Rajchl et al. [5]. It can be formulated as equation 1.

$$V_{new} = \frac{V_{old} - \mu}{\sigma} \quad (1)$$

where, $\mu, \sigma$ represent the mean and standard deviation of all voxels in the corresponding volume, respectively.

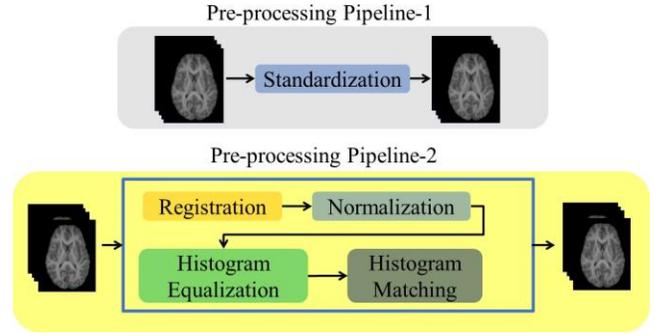

Fig. 2. Proposed pre-processing pipelines.

### B. Pre-processing Pipeline-2

*1) Registration:* It is the alignment of two or more images in a common anatomical space [12], commonly called the fixed space. As shown in Table I, training, validation and test subsets have different spacing. To standardize the voxel spacing in the dataset, we registered the images and transformed the labels correspondingly to Montreal Neurological Institute (MNI) template (152 subjects, 1x1x1mm T1-w, dimensions: 182 ×218 ×182, skull stripped) [13]. Fig. 3 shows the process of this registration. Simple-ITK framework in python was used for this purpose [14]. The idea is first to register the dataset to MNI template using rigid transformation, then save the corresponding final transformation matrix for transforming the labels. The available dataset was used as moving images, while, MNI template was used as the fixed image. After getting the final predicted labels, these should be brought back to the original space by applying the inverse of the corresponding saved transformation matrix.

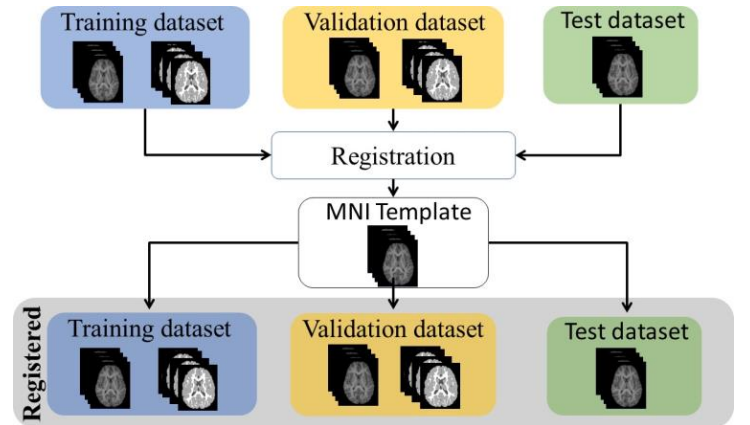

Fig. 3. Registering the whole dataset to MNI template to unify the voxel spacing.

*2) Normalization:* In this step, intensity rescaling was applied on each volume. The intensity range of each volume $v$ was rescaled to the range $[0, 1]$. It can be formulated as equation 2.

$$v_{new} = \frac{v - min(v)}{max(v) - min(v)} \quad (2)$$

*3) Histogram pre-processing:* In this step, the procedure was as follows:
  a) Reference selection: By analyzing tissues distribution for all volumes, image IBSR 07 was nominated due to the following reasons:
    - It has relatively wide spectrum of intensity values, see Fig. 4 left part.
    - The overlapping between GM and WM is acceptable, see Fig. 4 middle part.
    - GM and WM had comparable shares of voxels, this will help in the next step of histogram equalization of the reference image.
  b) Histogram equalization of the reference volume: adaptive histogram equalization was applied on the reference normalized volume IBSR 07. It's a process of adaptive image-contrast enhancement based on a generalization of histogram equalization (HE) [15].
  c) Histogram matching to the reference: apply histogram matching on all the dataset (minus IBSR 07) to the reference volume. As a last step of the pre-processing pipeline-2, histogram matching was performed to map all volumes histogram distributions to the reference volume's one. Fig. 5 shows 4 raw volumes and the corresponding pre-processed ones after applying pre-processing pipeline-2.

## IV. METHOD

### A. Network Architecture

In this paper, we adopted the NeuroNet [5] architecture for which the code is available at the Github repository of DLTK models [16]. NeuroNet is a deep convolutional neural network with multiple outputs, which was trained on 5,000 T1-weighted brain MRI scans from the UK Biobank Imaging Study that have been automatically segmented into brain tissue and cortical and sub-cortical structures using the standard neuroimaging pipelines [5]. For our work, as the desired output is tissue segmentation only, the architecture was modified to an updated FCN architecture [17] with a ResNet encoder [18] as presented in [19]. Fig. 6 shows the original NeuroNet architecture and the adopted architecture in this work. One initial convolution was performed on input volumes, afterwards, features were extracted using the ResNet encoder [18] [19]. Features were extracted in encoder part with two residual units ($N_{unit}$=number of residual units at each scale), $U_i^{S_j} = U_1^{S_j}, U_2^{S_j}$ on each of the resolution scales $S_j = \{S_1, ...., S_{N_{scales}}\}$, where $N_{scales}$ is the number of resolutions scales. We used $N_{scales} = 4$, as in the default implementation [5]. Leaky ReLu (with leakiness = 0.1) was used as the activation function [20] with preceding batch normalization. At each scale, the down-sampling was performed using stride convolution [21] where the strides were $s_j = 1, \{2, 2, 2\}$ operating on each spatial dimension. As defined in the reference paper [5], the number of filters for the convolutions in any $U^{S_j}$ was double the number at the previous scale: 16, 32, 64 and 128.

In image segmentation, Fully Convolutional Networks (FCNs) are among the widely-used networks which typically reconstruct the prediction with the same size of the input given. In the original Neuronet architecture, the decoding part was based on multi-decoder architecture on FCN upscore operations [17]. The prediction was reconstructed at each resolution scale $S_{j-1}$ by up-sampling the prediction linearly at $S_j$ scale and adding skip connection from the output of the last residual unit $U_2^{S_j}$. The output of the last residual unit at decoder serves as output of the network. A prediction was obtained after a softmax layer. The loss was calculated using categorical cross-entropy loss for all prediction outputs $\hat{y}$ at voxel locations $v$.

$$L(\hat{y}, y) = - \Sigma \hat{y}(v) \log y(v) \quad (3)$$

Where $y, \hat{y}$ are the true, predicted label, respectively.

## V. IMPLEMENTATION AND ENVIRONMENT

NeuroNet is implemented using Deep Learning Toolkit (DLTK) for Medical Image Analysis [19] on TensorFlow [22] with SimpleITK [23]. Pre-prpcessing pipelines used in this work were implemented using DLTK and SimpleITK frameworks as well. The training was carried out on GeForce GTX 1080 GPU with a memory of 2.7GB.

## VI. EXPERIMENTS AND RESULTS

Dice Similarity Coefficient (DSC) was used for evaluating the predicted labels and guiding the tuning process. Our main aim was to analyse the performance of the model with default parameters suggested in [5] with the proposed pre-processing pipelines and tune the parameters to improve the results. The corresponding results are shown in Table II.

Firstly, we trained our model with the processed data that went through only standardization (zero mean and unit standard deviation) see section III-A. The model was trained from scratch, but in the case of training deep models, enough care needs to be taken to initialize the parameters as shown in [24]. We used the pre-trained weights of Neuronet [5] as the initial ones to avoid gradient vanishing problem. We trained the model for 1000 steps with 200 randomly extracted patches of size $128 \times 128 \times 128$. The reason beyond choosing such a big patch size was that authors in [5] used it in the original implementation. Performance on the validation data is shown in Table II corresponding to model 1.1.

Secondly, the training steps were increased five times and the model was trained with double the number of patches (400 patches). That caused a huge improvement in the model's performance in CSF and a slight increase and decrease in WM and GM, respectively. Results are shown in Table II corresponding to model 1.2. Afterwards, more training was performed but no significant improvement was achieved, which led to looking for different pre-processing strategies.

Thirdly, the proposed pre-processing pipeline-2 was applied on the datasets explained in section III-B. The pipeline consists of registration, normalization, adaptive histogram equalization and histogram matching steps. We tried three different cases using pipeline-2 for pre-processing, these cases were as follows:

- Case 1 : patch size: 32x32x32, 200 samples.
- Case 2 : patch size: 64x64x64, 200 samples.
- Case 3 : patch size: 128x128x128, 50 samples.

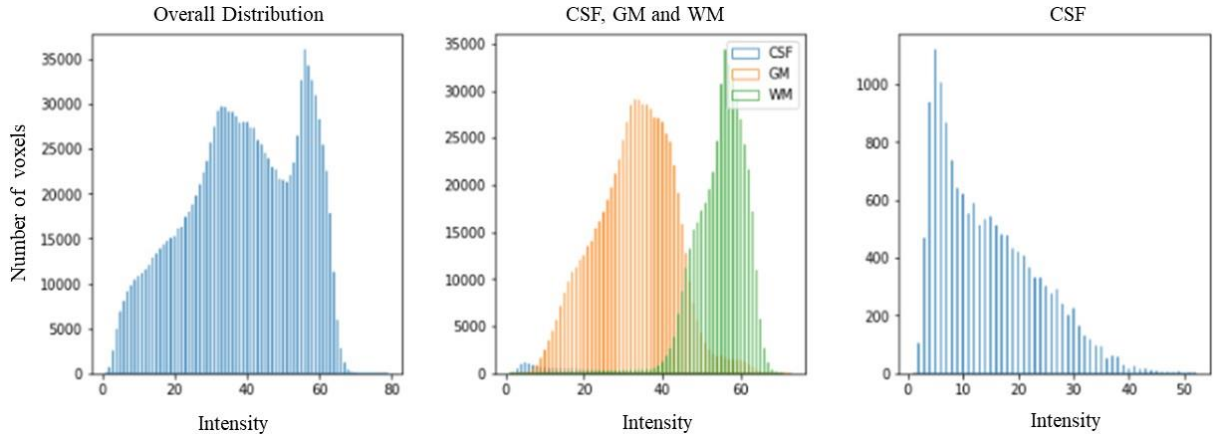

Fig. 4. IBSR 07 intensity distributions, where, on the left, the complete tissues intensity distribution, while, in the middle, CSF, white matter and GM distributions with different colours. The figure on the right is for CSF distribution only.

TABLE II
VALIDATION DSC OF DIFFERENT NETWORKS TRAINED WITH DIFFERENT HYPER-PARAMETERS. BOLD FONT HIGHLIGHTS THE HIGHEST DSC.

| Model Configuration | | | | | Dice Coefficient Validation Set | | |
|---|---|---|---|---|---|---|---|
| Model No. | #Training Steps | Patch Size | Samples | Weights Initializations | CSF | GM | WM |
| 1.1 | 1000 | 128x128x128 | 200 | NeuroNet Pretrained | 0.43±0.40 | 0.90±0.01 | 0.88±0.06 |
| 1.2 | 5000 | 128x128x128 | 400 | NeuroNet Pretrained | 0.80±0.12 | 0.89±0.05 | 0.89±0.03 |
| 2.1 | 4000 | 32x32x32 | 200 | Uniform Distribution | 0.07±0.03 | 0.71±0.04 | 0.72±0.06 |
| 2.2 | 4000 | 64x64x64 | 200 | Uniform Distribution | 0.80±0.05 | 0.90±0.02 | 0.89±0.03 |
| 2.3 | 4000 | 128x128x128 | 50 | Uniform Distribution | 0.89±0.02 | 0.93±0.01 | 0.93±0.01 |
| 2.4 | 2000 | 128x128x128 | 50 | NeuroNet Pretrained | 0.89±0.02 | 0.94±0.01 | 0.93±0.01 |
| **2.5** | **4000** | **128x128x128** | **50** | **NeuroNet Pretrained** | **0.90±0.02** | **0.94±0.01** | **0.93±0.02** |

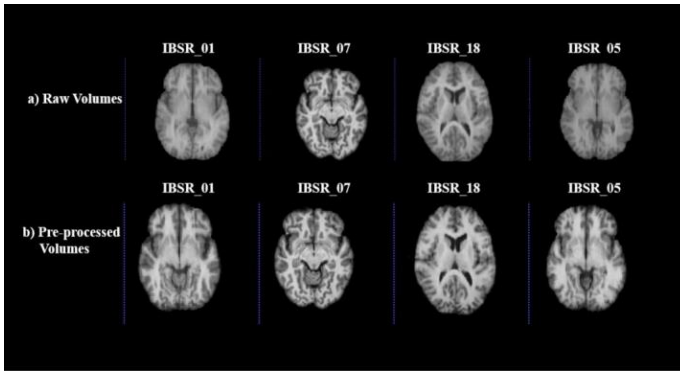

Fig. 5. The effect of pre-processing the dataset. At the top, 4 cases before applying the pre-processing pipelines. At the bottom, the final pre-processed dataset.

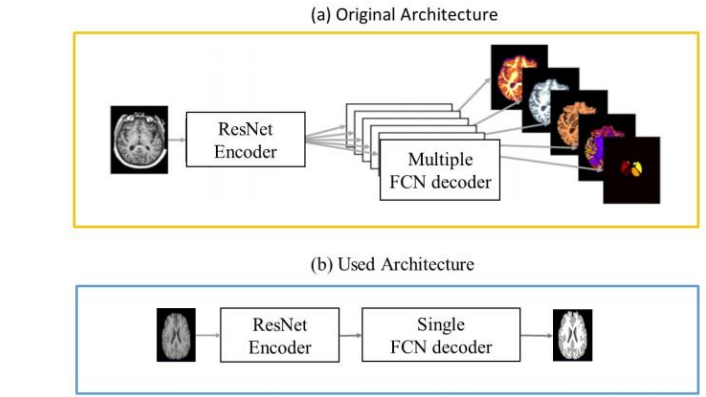

Fig. 6. Network Architectures [5].

Table II, model 2.1, 2.2 and 2.3 shows Case 1, Case 2 and Case 3 results respectively. Model 2.1, 2.2 and 2.3 were trained from end to end using the uniform distribution as weight initialization. Model 2.1, which was trained on patches of size 32x32x32, performed poorly and failed to segment CSF region. Training the model with bigger patch sizes turned out to be better as used in the original implementation. Overall, this model, 2.5, proved to have the best combination of parameters compared to all other experiments shown in Table II. Finally, we performed the test on the test set using the best model (2.5) and achieved average DSC 0.84 for CSF, 0.94 for GM and 0.94 for WM.

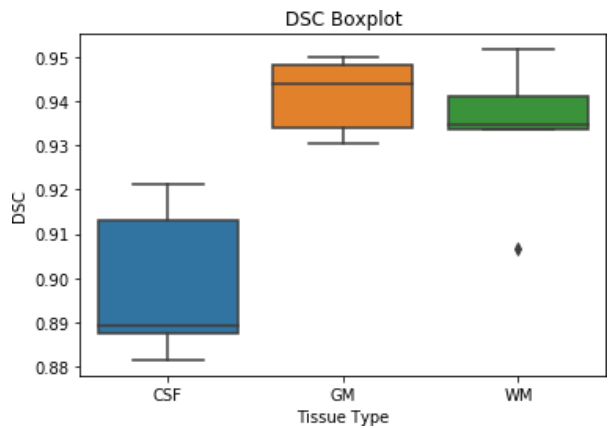

Fig. 7. DSC of best performing model 2.5 on the validation dataset.

## VII. Conclusion

In this work, we proposed using NeuroNet architecture for 3D brain tissue segmentation and investigated the performance of the CNN under different hyper-parameters and preprocessing techniques. To the best of our knowledge, this study is the first effort to use Neuronet with IBSR 2018 dataset by the time of this work. Our results show that histogram preprocessing was clearly effective in boosting the performance. Registration assisted mainly in unifying the voxel spacing to make all the inputs have similar spatial spaces. The use of NeuroNet pre-trained weights helped significantly the network to start from a good initial point. As compared to starting with random initialization, the pretrained weights were performing enormously better even though the size of the used dataset was relatively small (10 volumes compared to 5000 in [5]). Patch sizes had important effect on the performance as well. The original patch size was the best fit, while, smaller sizes just did not work as good. Future work includes extending the analysis with 3D Unet and modified ResNet-Unet architectures. We also plan to ensemble different models to improve the DSC for CSF especially. We also plan to extend the use of the same CNN in disease segmentation.